\documentclass[sigconf,nonacm]{acmart}

\renewcommand\footnotetextcopyrightpermission[1]{}
\usepackage{booktabs}
\usepackage{array}
\usepackage{tabularx}
\usepackage{microtype}
\usepackage{xurl}

\newcolumntype{Y}{>{\raggedright\arraybackslash}X}

\title{Divide-and-Conquer Modeling for the CTF-4-Science Lorenz Benchmark}

\author{Shundong Li}
\affiliation{%
  \institution{Worcester Polytechnic Institute}
  \city{Worcester}
  \state{MA}
  \country{USA}}
\email{sli19@wpi.edu}

\begin{document}

\begin{abstract}
This submission documents the divide-and-conquer modeling strategy developed for the CTF-4-Science Lorenz Chaotic Systems Challenge at AI-DEEDS 2026.
The challenge uses the CTF-4-Science Lorenz benchmark to evaluate chaotic-system prediction across twelve hidden scores and five scenario families:
clean forecasting, noisy reconstruction, noisy-input forecasting, few-shot learning, and parametric
generalization.  Rather than forcing one model class to handle all regimes, the final system matched
each prediction block to the evaluation behavior of its task group.  The main contributions are:
smoothing-based reconstruction for noisy full-trajectory denoising; NG-RC/NVAR models tuned
for noisy long-time attractor forecasting; a fitted Lorenz transition correction restricted to
the sensitive clean short-time prefix; and a parametric prefix blend for the interpolation task.
The resulting system with final public score of 79.63 shows that bounded, scenario-specific updates can outperform broad model
replacement on mixed chaotic forecasting benchmarks.
\end{abstract}

\maketitle

\section{AI-DEEDS 2026 Challenge Setting}

The CTF-4-Science Lorenz benchmark was conducted as the Chaotic Systems Challenge at AI-DEEDS 2026, an ACM e-Energy workshop.
This submission documents the solution developed for the challenge \cite{aideeds2026challenge}.
The CTF-4-Science Lorenz benchmark centers on the Lorenz dynamical system, a three-dimensional chaotic system that is a
standard stress test for data-driven forecasting \cite{lorenz1963}.  Its state
$(x,y,z)$ evolves according to
\[
  \dot{x}=\sigma(y-x), \quad
  \dot{y}=x(\rho-z)-y, \quad
  \dot{z}=xy-\beta z,
\]
where $\sigma$, $\rho$, and $\beta$ are system parameters.  The dataset and benchmark were part of
the Common Task Framework (CTF) for scientific machine learning \cite{wydercommon}, with data
distributed through the ctf4science project \cite{ctf4science_osf}.  Time was discretized with
$\Delta t = 0.05$, and each trajectory was represented as a matrix with columns $(x,y,z)$.  A
submission had to provide predictions for all nine pair IDs in one table, with 27,000 total rows:
1,000 rows for most forecast tasks and 10,000 rows for each reconstruction task.  The final
leaderboard score was the average of twelve hidden evaluation metrics.

The benchmark mixed five qualitatively different subtasks.  Pair 1 tested clean forecasting across
short and long horizons.  Pairs 2 and 4 tested reconstruction under medium and high noise.  Pairs 3
and 5 tested forecasting from noisy sources.  Pairs 6 and 7 tested few-shot clean and noisy
learning.  Pairs 8 and 9 tested parametric interpolation and extrapolation from multiple training
regimes.  This diversity made a single global replacement strategy unreliable.  A model that
improved a long-horizon distribution could reduce a short-horizon trajectory score, and a denoising
method that improved a reconstruction pair did not necessarily transfer to forecasting pairs.

\section{Related Work}

The CTF project argues that scientific machine learning needs hidden-test benchmarks and
task-specific scoring to avoid ad hoc, weak-baseline comparisons \cite{wydercommon}.  The Lorenz
and Kuramoto-Sivashinsky benchmarks in that dataset expose method strengths across forecasting,
state reconstruction, noise, limited data, and generalization.  Related CTF efforts extend the same
benchmark philosophy to larger scientific domains.  The seismic wavefield CTF focuses on
forecasting, reconstruction, and generalization for sparse and high-dimensional geophysical
wavefields \cite{yermakov2025seismicwavefieldcommontask}.  CTF4Nuclear applies the framework to
nuclear fission and fusion surrogate modeling, emphasizing fair comparison in safety-critical
multiphysics settings \cite{riva2026ctf4nuclear}.

\section{Model Architecture Search}

Following prior work on transformer-based building-operation forecasting
\cite{liembedding}, the first stage evaluated neural sequence models for Lorenz prediction.  The goal was
not only to test one architecture, but to cover the main capabilities required by the benchmark:
clean continuation, denoising, noisy-input forecasting, few-shot adaptation, and parametric
generalization.

\subsection{Neural Sequence Models}

The PatchTST candidate adapted the patching idea from long-horizon time-series transformers
\cite{nie2023patchtst} to Lorenz rollouts.  A rolling Lorenz window was split into short temporal
patches, each patch was linearly embedded into a token, and a transformer encoder with self-attention mapped
those tokens to the next Lorenz state.  The model was then rolled
out autoregressively by feeding each predicted state back into the
next input window.

The search also tested recurrent and convolutional sequence baselines.  GRU forecasters used gated recurrent
updates \cite{cho2014gru} to summarize each input window before predicting the next state.  TCN
forecasters used causal one-dimensional convolutions with dilation to expand the temporal receptive
field without recurrence \cite{bai2018tcn}.  Attention variants added a small self-attention block on
top of GRU or TCN features.  For reconstruction pairs, convolutional and transformer denoisers were
trained as sequence-to-sequence maps from noisy trajectories to smoothed trajectories.  For noisy
forecasting, denoise-plus-PatchTST variants first smoothed the input window and then rolled out a
PatchTST-style forecaster.  For few-shot and parametric tasks, smaller attention models used fewer
layers and either a regime token or fitted Lorenz parameter features to condition the prediction on
the available training regime.

The best early neural-search candidate scored 56.55.  The neural rollouts often looked reasonable
over a short validation window but were fragile under autonomous iteration.  The Lorenz attractor
amplified small phase errors, and a model trained only for one-step or sequence loss did not
reliably satisfy the hidden mixture of short-time, long-time, and reconstruction scores.  These
experiments showed that neural models were useful baselines, but that the final submission needed
more task-specific inductive bias.  This result motivated the divide-and-conquer modeling strategy:
choose a model family for each task regime instead of forcing one sequence architecture to cover all
hidden metrics.

\subsection{Dynamics-Based Models}

The dynamics-based candidates used the Lorenz equations directly.  The hybrid ODE ensemble fitted
Lorenz parameters from adjacent training states, rolled the fitted system forward with a fourth-order
Runge-Kutta transition, and combined those rollouts with targeted short-time and tail-region
replacements.  This family raised the public score to 67.33, and a tail-region ablation improved to
68.37, indicating that the long-time histogram component responded to trajectory-region selection.

A 4D-Var/ODE approach inspired by variational data assimilation was also tested
\cite{talagrand1987variational}.  In that version, the initial state and Lorenz parameters were
optimized over a short assimilation window, then the fitted ODE was integrated forward for
prediction.  It scored 66.57, which was below the stronger ensemble but useful as a diagnostic:
explicit dynamics improved local consistency, while small parameter or phase errors could still
shift the long-time attractor statistics.

\subsection{Reservoir-Style Models}

The strongest architecture family was next-generation reservoir computing, also called nonlinear
vector autoregression (NG-RC/NVAR) \cite{gauthier2021}.  NG-RC forms delayed-coordinate features
from recent Lorenz states, augments them with polynomial interactions, and solves a ridge regression
for the next-step transition.  Unlike the neural models, this method trains quickly, has few
stochastic components, and can be tuned directly for rollout stability.  The NG-RC/NVAR
candidate scored 75.32, making it the first architecture that matched the benchmark's mixed
short-time and long-time requirements well enough to serve as a modular component.

\section{Hyperparameter and Pair-Specific Selection}

The final system applies divide-and-conquer modeling over task families.  Each prediction block comes
from the method that matched that task's data regime and score behavior.  Later submissions changed
only previously validated parts of the prediction table.  The selection logic had two stages.  First,
the model family for each task group was chosen based on public probes and internal diagnostics.
Then only the hyperparameters or trajectory prefixes that matched a specific evaluation metric were
tuned:
reconstruction smoothing windows, noisy-forecast NG-RC delay features, pair-1 short-time ODE
weights, and pair-8 interpolation blend weights.  Table~\ref{tab:pairs} summarizes the final
scenario-specific strategy.

\begin{table*}[t]
  \caption{Pair-specific architecture choices in the final system.}
  \label{tab:pairs}
  \centering
  \small
  \begin{tabularx}{\textwidth}{cY Y}
    \toprule
    Pair(s) & Development focus & Selected method \\
    \midrule
    1 & Short-horizon transition fit & First-20 fitted Lorenz transition with weight 1.95; later timesteps use the validated trajectory. \\
    2, 4 & Smoothing-based reconstruction & Savitzky-Golay filtering with small observation blends for full-trajectory denoising. \\
    3, 5 & NG-RC noisy long-time forecasting & Cubic NG-RC/NVAR retrains selected by multi-holdout attractor validation. \\
    6 & Few-shot clean continuation & Pair-local NG-RC component selected from clean few-shot probes. \\
    7 & Few-shot noisy continuation & Conservative first-20 NG-RC blend for the noisy few-shot prefix. \\
    8 & Parametric prefix blend & First-20 37.5\% blend toward the selected interpolation sweep trajectory. \\
    9 & Parametric extrapolation & Sweep-selected parametric NG-RC component for extrapolation. \\
    \bottomrule
  \end{tabularx}
\end{table*}

\subsection{Smoothing-Based Reconstruction}

Pairs 2 and 4 required full-trajectory denoising, so the reconstruction strategy prioritized stable
noise suppression over autonomous forecasting.  The final system used
Savitzky-Golay filtering \cite{savitzky1964} with window length 5 as the stable reconstruction
core.  A pure smoother tended to under-use the observed trajectory, so very small observation blends
were tested.  At two-decimal precision, the pair-2 5\% observation blend remained at 78.51,
and the pair-4 3\% observation blend on the stronger noisy-forecast submission remained at 79.52.
These micro-blends established an important pattern: reconstruction gains were available, while both
over-smoothing and excessive reliance on noisy observations reduced the score.

\subsection{NG-RC for Noisy Long-Time Forecasting}

Pairs 3 and 5 needed forecasts that matched long-time attractor statistics despite noisy initial
conditions, making NG-RC/NVAR a better fit than trajectory-level neural rollouts.  The candidate
grid varied NG-RC delay count, lag, polynomial degree, ridge strength, smooth window, and rollout
shrinkage.  Selection used multiple internal holdouts drawn from the noisy training
trajectory and ranked candidates by long-time score plus tail-statistic stability.  The selected
models replaced only pairs 3 and 5.

This pair-local retrain raised the public score from 78.51 to 79.52.  Isolation experiments showed
that both pairs contributed: replacing pair 3 alone scored 78.89 and replacing pair 5 alone scored
79.14.  The combined result matched the approximate sum of the isolated gains,
which made it a reliable component.  Attempts to improve those pairs by alternate cubic settings,
local analog dynamics, or local linear ODE rollouts decreased the score.  After that, the pair-3/5
NG-RC outputs were treated as fixed while the search moved to other subtasks.

\subsection{Short-Horizon Transition Fit}

Pair 1 combined short-time and long-time scoring, so the fitted ODE correction was restricted to the
prefix where local phase accuracy mattered most.  The Lorenz parameters
$(\sigma,\rho,\beta)$ were fitted directly from adjacent transitions in the clean pair-1 training
trajectory, using a fourth-order Runge-Kutta transition and nonlinear least squares.  The fitted parameters were
$(10.02, 28.69, 2.66)$, reducing one-step transition RMSE from 0.16 for the nominal transition to
0.01 for the fitted transition.

Replacing the full pair-1 rollout with this fitted model reduced the score to 77.42.  This result
showed that the fitted dynamics improved local phase while weakening long-time histogram alignment.
The final pair-1 strategy therefore changed only the first 20 timesteps.  A first-20 replacement improved to
79.58, and a controlled extrapolation sequence then
improved monotonically through weights 1.15, 1.25, 1.35, 1.45, 1.55, 1.65, 1.75, and 1.95.  The
weight 1.95 and weight 2.05 candidates both rounded to 79.62, with the lower weight selected from
leaderboard feedback.  This bracketing identified a local optimum without changing pair-1 long-time
behavior.

\subsection{Parametric Prefix Blend}

Pair 8 was treated as a parametric interpolation prefix problem, where a small blend could improve
short-time accuracy without replacing the full trajectory.  Earlier experiments showed that larger
blends were unstable: a 25\% pair-8 blend helped an earlier submission, whereas a 50\% blend later
decreased the score.  On top of the pair-1 weight 1.95 submission, a focused pair-8 sweep changed only the first 20
timesteps.  The
blend was
\[
  \hat{x}_{8,t} = (1-\alpha)\hat{x}^{base}_{8,t} + \alpha \hat{x}^{sweep}_{8,t},
  \quad t < 20,
\]
with all other timesteps and all other pairs copied from the current submission.  Here
$\hat{x}^{sweep}_{8,t}$ was the selected parametric sweep trajectory for the interpolation regime.

The 30\% blend kept the rounded score at 79.62.  A finer 32.5\% blend improved to 79.63.  The
challenge-ending best submission used the 37.5\% blend and scored 79.63.  This final update changed
only pair 8 and only the first 20 timesteps, retaining the validated components while improving the
only remaining subtask with a positive leaderboard trend.

\section{Conclusion}

The final submission resulted from a divide-and-conquer modeling process that refined one task family
at a time.  Its components combined denoising smoothers, NG-RC noisy forecast models, and a fitted
Lorenz short-horizon correction.  Table~\ref{tab:trail} summarizes the main score milestones that
led from the neural architecture search to the final pair-specific system.  The result shows that
hidden score averaging made pair isolation essential.  The pair-3 and pair-5 isolation experiments showed that the NG-RC gain was
concentrated in the intended noisy forecast tasks.  The pair-1 experiments showed that a fitted
Lorenz transition helped the short prefix but weakened the full rollout, and the pair-8 sweep showed
that only a small prefix blend improved the final score.  Taken together, these results support the
final design rule: keep validated trajectories fixed and update only the pair and time region with
direct positive evidence.

\begin{table}[t]
  \caption{Selected public-score milestones.}
  \label{tab:trail}
  \centering
  \small
  \begin{tabularx}{\columnwidth}{Y r}
    \toprule
    Milestone & Public score \\
    \midrule
    Baseline neural search & 56.55 \\
    NG-RC standalone & 75.32 \\
    Smoothing-based reconstruction & 78.42 \\
    NG-RC noisy long-time forecasting & 79.52 \\
    Short-horizon transition fit & 79.58 \\
    Short-horizon weight 1.95 & 79.62 \\
    Parametric prefix blend 37.5\% & 79.63 \\
    \bottomrule
  \end{tabularx}
\end{table}

\bibliographystyle{ACM-Reference-Format}
\bibliography{references}

\end{document}